\documentclass[conference]{IEEEtran}
\IEEEoverridecommandlockouts
\usepackage{cite}
\usepackage{amsmath,amssymb,amsfonts}
\usepackage{algorithmic}
\usepackage{graphicx}
\usepackage{textcomp}
\usepackage{xcolor}
\def\BibTeX{{\rm B\kern-.05em{\sc i\kern-.025em b}\kern-.08em
    T\kern-.1667em\lower.7ex\hbox{E}\kern-.125emX}}

\usepackage{subcaption} 
\usepackage[inline]{enumitem}
\usepackage{xcolor}
\usepackage{epsfig}
\usepackage{algorithm}
\usepackage{cleveref}
\usepackage{booktabs}
\usepackage{multirow}
\usepackage{tikz}
\usepackage{comment}
\usepackage{colortbl}
\usepackage{lipsum}
\usepackage{balance}
\usepackage{bm}
\usepackage{dsfont}

\newtheorem{oss}{Observation}
\DeclareMathOperator*{\argmin}{argmin} % thin space,

\begin{document}

\title{Attention-Based Real-Time Defenses for Physical Adversarial Attacks in Vision Applications}

% \author{\IEEEauthorblockN{Giulio Rossolini}
% \IEEEauthorblockA{\textit{Dept. of Excellence in Robotics \& AI} \\
% \textit{Scuola Superiore Sant'Anna}\\
% Pisa, Italy\\
% giulio.rossolini@santannapisa.it}
% \and
% \IEEEauthorblockN{Alessandro Biondi}
% \IEEEauthorblockA{\textit{Dept. of Excellence in Robotics \& AI} \\
% \textit{Scuola Superiore Sant'Anna}\\
% Pisa, Italy\\
% alessandro.biondi@santannapisa.it}
% \and
% \IEEEauthorblockN{Giorgio Buttazzo}
% \IEEEauthorblockA{\textit{Dept. of Excellence in Robotics \& AI} \\
% \textit{Scuola Superiore Sant'Anna}\\
% Pisa, Italy\\
% giorgio.buttazzo@santannapisa.it}
% }

\author{
Giulio Rossolini,~Alessandro Biondi, ~and Giorgio Buttazzo
\\
\IEEEauthorblockA{Department of Excellence in Robotics \& AI, Scuola Superiore Sant'Anna, Pisa, Italy \\ name.surname@santannapisa.it}
\thanks{This work has been submitted to the IEEE for possible publication. Copyright may be transferred without notice, after which this version may no longer be accessible.}
}

% \author{\IEEEauthorblockN{Anonymous Authors}
% email address or ORCID}}


\maketitle
\begin{abstract}  
Deep neural networks exhibit excellent performance in computer vision tasks, but their vulnerability to real-world adversarial attacks, achieved through physical objects that can corrupt their predictions, raises serious security concerns for their application in safety-critical domains. Existing defense methods focus on single-frame analysis and are characterized by high computational costs that limit their applicability in multi-frame scenarios, where real-time decisions are crucial.

To address this problem, this paper proposes an efficient attention-based defense mechanism that exploits adversarial channel-attention to quickly identify and track malicious objects in shallow network layers and mask their adversarial effects in a multi-frame setting.
This work advances the state of the art by enhancing existing over-activation techniques for real-world adversarial attacks to make them usable in real-time applications. It also introduces an efficient multi-frame defense framework, validating its efficacy through extensive experiments aimed at evaluating both defense performance and computational cost.
\end{abstract}



%\keywords{adversarial attacks, real-world adversarial defense, neural network analysis, robust and secure AI}

\maketitle

\section{Introduction}
\label{sec:intro}
In recent years, \emph{deep neural networks} (DNNs) have demonstrated remarkable performance in several computer vision tasks. At the same time, they have been shown to be quite vulnerable to adversarial attacks~\cite{Szegedy14}, where small perturbations of input data can cause a model to output wrong predictions.
To address this problem, an increased research effort has been devoted to make DNNs more reliable, robust, and secure, to be adopted in \emph{cyber-physical systems} (CPS), as autonomous vehicles and robots~\cite{shea2021algorithmic, mohan2013s3a, sun_survey_2018}.

%---------------------------------------------------
% METTERLE DA QUALCHE PARTE
%\cite{kantaros2021real, zhou2022robustness}

Although adversarial perturbations represent a concrete security threat for DNNs, they raised significant discussions in the CPS community, mainly questioning the practical relevance of these attacks. It is indeed not entirely realistic to consider threat models in which the attacker has access to the digital representation of the frames captured by a vision system, to run adversarial attacks against DNNs, while not having the capability of compromising other software components in the system that could be even easier to attack. 
In response to this argument, research efforts have been shifted towards \textit{real-world adversarial attacks}~\cite{pmlr-v80-athalye18b}, which are deployed through \emph{physical} objects, such as billboards and patches, that are specifically crafted and strategically placed in the external environment to fool DNNs~\cite{rossolini_tnnls_2023, wang2022survey}.

\begin{figure*}[ht]
\begin{center}
\includegraphics[width=\textwidth]{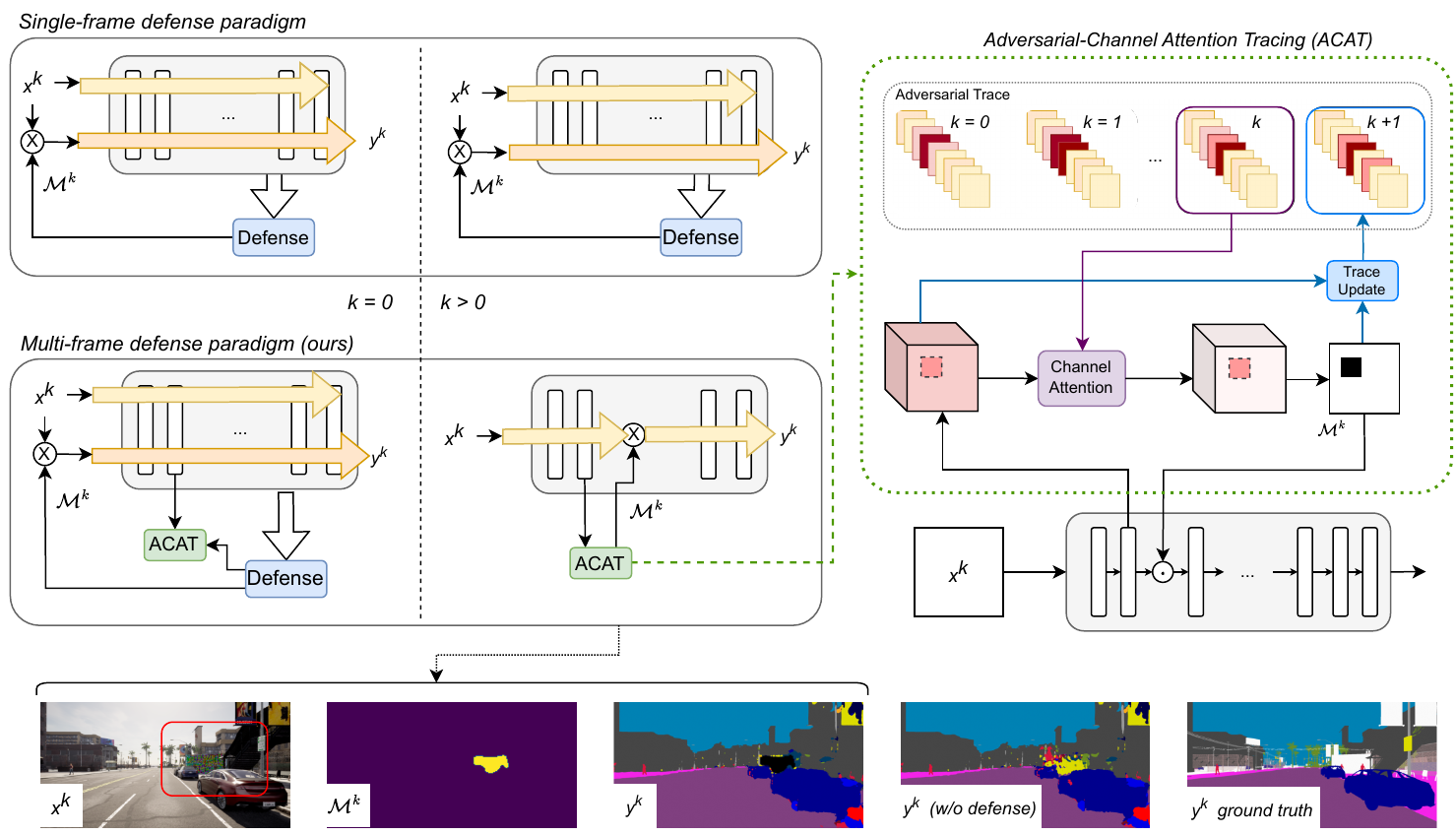}
\caption{\small{
Schematic and simplified overview of the proposed multi-frame defense paradigm compared to state-of-the-art, single-frame defense paradigms. At frame $k=0$, with a first inference pass (yellow arrow in the figure), a single-frame defense mechanism extracts a mask to inhibit the detected attack. Another inference pass is required at frame $k=0$ to apply the defense mask (orange arrow in the figure). With the proposed approach, the mask is used to implement pattern analysis in the shallow layers for the next frames $k>0$, which is the core task performed by ACAT. This allows extracting an \emph{adversarial trace} that allows for a quick identification of the shape of the adversarial object, hence efficiently generating and applying defense masks (right side of the figure). At the bottom, we show illustrations of the defense mechanism in a simulated attacked Carla driving scenario \cite{nesti2022carla} with the BiseNet model \cite{bisenet_paper}, where the adversarial object is highlighted in the red area. For completeness, we also report the output of the same frame without any defense mechanism and the ground truth.}
}
\label{f:fig_intro}
\end{center}
\end{figure*}

To enhance the robustness of DNNs against such real-world adversarial attacks, various techniques have been proposed in the literature (discussed in Section~\ref{sec:rel}). A common paradigm that can be found in previous work consists in producing at run-time a mask to cover adversarial objects, thereby preserving the predictions of the DNNs under attack.

%{\color{blue} 
Although recent defense methods have shown a promising performance to contrast such types of attacks, even on complex real-world scenarios, previous work mainly focused on single-image (i.e., single-frame) cases and without paying particular attention at the computational cost of the proposed defense method, resulting in more inference passes or additional expensive neural models.
These limitations make state-of-the-art approaches inadequate for CPS, where efficient solutions capable to operate in real time on video streams (i.e., multiple frames) are required.

\smallskip
\noindent
\textbf{This work.}
To face these challenges, we take inspiration from recent studies \cite{wu_defending_2020, rossolini2023defending} that assess strong and provable connections between anomalous over-activations in deep network layers and real-world adversarial effects on the model output. 
In particular, this work delves deep into understanding the over-activation phenomenon by observing the presence of specific channels even in the first layers of DNNs, which are predominantly targeted by the real-world attack for propagating adversarial effects. We systematically identified this attack pattern through channel-wise weights, denoted as \textit{adversarial trace}, that enable a significantly faster and more accurate identification of attacks by means of a proper attention strategy designed in this work. This allows for the immediate removal of adversarial features before their spatial propagation in the deep layers, hence detecting and masking attacks in a single inference pass.
%}

After presenting the results of our analysis and providing insights into the nature of the adversarial trace, we propose a defense algorithm for multi-frame vision applications named \textit{Adversarial-Channel Attention Tracing} (ACAT). To enable the efficient tracing of adversarial physical objects in a video stream, ACAT requires to know a starting spatial position of the objects, which can be extracted using a single inference pass of state-of-the-art single-frame defense methods. As witnessed by the experimental results reported in the paper, this improves both efficiency in terms of running times and computing load, as well as the attack detection effectiveness.
The proposed approach is illustrated in Figure~\ref{f:fig_intro}.
%shows an illustration of the above-mentioned observations and schematizes the proposed approach.

In summary, this work makes the following contribution:
\begin{itemize}
\item It advances the understanding of adversarial over-activations in shallow DNN layers when aiming at detecting real-world adversarial attacks, hence introducing the concept of the \textit{adversarial trace}.
\item It proposes \textit{ACAT}, an algorithm for multi-frame applications based on a channel-attention mechanism to make more computationally efficient and more effective the defense from real-world adversarial attacks.
\item It presents extensive experiments and ablation studies to show the benefits of the proposed approach in terms of defense performance and computational costs, focusing on autonomous driving scenarios.
\end{itemize}

%###############################################################
% - citazioni ICCPS
%###############################################################
% A-https://scholar.google.com/citations?view_op=view_citation&hl=en&user=rBoCJHwAAAAJ&sortby=pubdate&citation_for_view=rBoCJHwAAAAJ:fPk4N6BV_jEC
% B- https://scholar.google.com/citations?view_op=view_citation&hl=en&user=Jbo1MqwAAAAJ&citation_for_view=Jbo1MqwAAAAJ:tOudhMTPpwUC
% C-(almeno un altro - forse il survey di y.sun???)

%D-https://scholar.google.com/citations?view_op=view_citation&hl=en&user=qPlUgrgAAAAJ&cstart=20&pagesize=80&sortby=pubdate&citation_for_view=qPlUgrgAAAAJ:R6WN2b6jgFYC
%E-https://scholar.google.com/citations?view_op=view_citation&hl=en&user=B_ouhTgAAAAJ&sortby=pubdate&citation_for_view=B_ouhTgAAAAJ:oNZyr7d5Mn4C
%F- https://scholar.google.com/citations?view_op=view_citation&hl=en&user=sXpmLxUAAAAJ&sortby=pubdate&citation_for_view=sXpmLxUAAAAJ:VaXvl8Fpj5cC
%G - survey Y.Sun

%################################################# Others...

\section{Related Work}
\label{sec:rel}
\paragraph{Real-world adversarial attacks}
In the context of the analysis of adversarial perturbations \cite{kurakin_adversarial_2017}, real-world (RW) attacks have received particular interest from the secure AI community, due to their capability of fooling the model outcomes from the physical environment in which they operate. Indeed, from the standpoint of the attacker, the RW attack paradigm ideally avoids injecting adversarial features digitally, thus circumventing the need for compromising a computing system.
%{\color{blue}
To this end, different use cases addressed in the literature illustrate how physical attacks pose significant threats to AI systems. These include deceiving intrusion detection systems \cite{wu2020making, hu2022adversarial, adversarialtshirt_Xu_2020}, manipulating the identification of pedestrians or cars in driving scenarios \cite{rossolini_tnnls_2023, zhang_camou_2019}, and fooling steering angle predictor \cite{wu_physical_2020}.
From an architectural point of view, all the vision models can be susceptible to physical attacks, as image classification \cite{brown_adversarial_2018, pmlr-v80-athalye18b}, semantic segmentation \cite{rossolini_tnnls_2023}, object detection \cite{patch_object}, depth estimation \cite{cheng2022physical}.

To comprehensively assess the model's robustness against these threats, recent studies have also emphasized the necessity of proposing proper benchmarks to evaluate model robustness against RW attacks \cite{nesti2022carla, braunegg2020apricot}.
%}

\paragraph{Defense methods}
To enhance the robustness of vision models against these attacks, various defense mechanisms have been proposed. While some focus on flagging the presence of attacks only, thus allowing to just reject the attacked frames \cite{rossolini_tnnls_2023, co_real,xiang_patchguard2_2021, xiang_patchcleanser_2021, chou_sentinet_2020}, more sophisticated mechanisms aimed at mitigating the attack effects at run-time, providing an attack-free DNN output. The main idea involves segmenting the position of the adversarial object within the image with the purpose of generating a pixel mask, which is capable of inhibiting the attack effects in the input space or directly in the network layer. 
%removing it to retrieve an not-adversarial outcome.

Some techniques \cite{liu2021segment, chiang_adversarial_2021, xu2023patchzero} used a secondary encoder-decoder model to compute the mask. The mask is then used to eliminate the adversarial attack from the image before it is passed to the DNN of the target vision application.
These approaches significantly increase the overall computational cost for each input (even for the non-attacked ones).
More classic approaches instead, as LGS \cite{naseer_local_2019}, aim at filtering out adversarial features from the image using gradient-based filters, assuming the adversarial features of objects have high frequency. 
%and are difficult to certify, as they rely on a secondary DNN.

Other strategies, instead, are based on internal analysis of DNN layers to identify anomalous over-activations at run time~\cite{wu_defending_2020, rossolini2023defending}, which proved to be highly correlated with real-world adversarial effects in any targeted vision tasks.
Specifically, these defense mechanisms extract masks by addressing the spatial over-activation of deep features. These approaches tend to exhibit a more predictable and robust behavior compared to those based on a secondary model.
Nevertheless, they require two inference (i.e. forward) passes as the attacks are detected when analyzing deep layers of the model, where the effects of the attacks are not anymore recoverable. The second pass is hence needed to process the input image with the pixel mask applied.

Overall, although all the approaches presented in previous work to defend from RW attacks have shown promising performance and the capability to generalize among different vision tasks, little to no efforts have been made 
in addressing their usage in real time within CPS applications.
%to address practical aspects crucial for concrete use in safety-critical systems, such as computational costs and predictability.

% GR - Toglierei la parte dopo, è già ripetuta nell'intro
% \paragraph{This work.}
% Inspired by the predictability of the relationship between over-activation and real-world (RW) attacks, this work aims to uncover an understanding of adversarial features in the first model stages, enabling masking directly at runtime without the need for a second inference. The main idea of our analysis is that an adversarial object can manipulate a limited set of representations in the input space, causing over-activation of deep layers. We assert that such deep over-activation is driven by specific features in the shallow layers, which are also over-activated. With this knowledge, it is possible to perform channel attention \cite{hu2018squeeze} on such patterns to promptly extract an adversarial mask.
% We noticed this simple principle aligns with a multi-frame segmentation tracking scenario, where, in the first attacked frame, a single-frame expensive defense method provides a reference position of the adversarial object. 
%, we can trace the adversarial object in the shallow layers, enhancing both performance and computational efficiency while keeping the robustness property inherited from the over-activation analysis.

\section{Background and Preliminaries}

This section concisely provides background and preliminary concepts for the rest of the paper.
We consider vision models that take as input an image with dimensions \(H \times W\) pixels and \(C\) channels, denoted by \(\mathbf{x} \in \mathbb{R}^{3 \times H \times W}\). The model output, denoted as \(f(\mathbf{x})\), depends on the specific vision task under consideration.

For simplicity, the notation is introduced by referring to a simple feed-forward DNN, with a list of layers $\{L^1, ..., L^{N_L}\}$, where the input is forwarded sequentially. To this end, we use the operational notation \(f^{i \to j}\) to denote the processing flow of features from layer \(L^i\) to layer \(L^j\). The layer index 0 is used to refer to the input of the model. For instance, $f(\mathbf{x})$ is equivalent to $f^{0 \to N_L}(\mathbf{x})$ or $f^{j \to N_L}(f^{0 \to j}(\mathbf{x}))$. 
%This allows to split the inference process in a intermediate layer. 
% AB - {\color{red} TODO: introduce here the \emph{activations} of a layer, also defining the corresponding symbol.} 
% GR - {\color{blue} migliorata definition in 3.2}

\subsection{Real-World Adversarial Attacks}
Real-world adversarial attacks can be generated by introducing \textit{adversarial physical objects} into specific regions of the scene captured by the input image \(\mathbf{x}\). Following previous work, we can model these objects as rectangular patches denoted by \(\bm{\delta}\), where \(\bm{\delta}\) is an image of size \(\tilde{H} \times \tilde{W}\) with \(C\) channels, where \(\tilde{H} \leq H\) and \(\tilde{W} \leq W\). Crafting an adversarial patch involves solving an optimization problem aimed at minimizing a specific attack loss function, while enhancing the robustness of the patch features against realistic transformations that can occur while filming the patch in the physical world~\cite{pmlr-v80-athalye18b}.

Formally, we craft an adversarial patch \(\bm{\delta}\) by solving the following optimization problem:
\begin{equation}
    \bm{\delta} = \argmin_{\bm{\delta}}~\mathbb{E}_{\mathbf{x}\sim\mathbf{X}, \gamma\sim\bm{\Gamma}} ~ \mathcal{L}_{\text{Adv}}(f(\tilde{\bm{x}}), \mathbf{y}_{\text{Adv}}),
\label{eq:adv}
\end{equation}

%AB - {\color{red} $\gamma$ dovrebbe essere usata da qualche parte nella formula}
%GR - Ok - dovrei aver sistemato

\noindent where \(\mathbf{X}\) is a set of images to attack, $\tilde{\bm{x}} = \bm{x} + \gamma(\bm{\delta})$ is the attacked image, \(\bm{\Gamma}\) is a set of appearance and placement transformations that can be randomly selected to apply a patch, 
\(\mathbf{y}_{\text{Adv}}\) is the adversarial output target, and \(\mathcal{L}_{\text{Adv}}\) is the adversarial loss function that specifies the objective of the attacker, the lower \(\mathcal{L}_{\text{Adv}}\)  the more adversarial effect.
% AB - {\color{red} Add here the key property of \(\mathcal{L}_{\text{Adv}}\): the lower the more adversarial effect (right?)}
% GR - Ho anche tolto la parte sotto per snellire. -->
%For instance, if the objective is to execute an untargeted attack, the adversarial target \(\mathbf{y}_{\text{Adv}}\) can be any but the ground truth \(\mathbf{y}\) and hence a classical cross-entropy loss function $\mathcal{L}$ can be used as \(-\mathcal{L}(f(\tilde{\mathbf{x}}), \mathbf{y})\).
Please refer to \cite{pmlr-v80-athalye18b, brown_adversarial_2018, rossolini_tnnls_2023} for further details.

\subsection{Defense Mechanisms and Internal Analysis}
\label{sec:internal_analysis}
As discussed in Section 2, several defense strategies have been proposed in the literature to mitigate real-world adversarial attacks, particularly in single-frame applications. In this context, our approach aligns closely with works that perform internal analysis of neural models during inference. 

Following this paradigm, we denote by $\bm{h}^{l} \in \mathbb{R}^{C^l \times H^l \times W^l}$ the features produced by any layer $L^l$, where $C^l, H^l, W^l$ are the corresponding tensor dimensions, i.e., $\bm{h}^{l}=f^{0 \to l}(\mathbf{x})$.
The notation $(T)_{c, i, j}$ is used to denote a single element of any 3D tensor $T$, where $c$, $i$, and $j$ are the indices for the channel, height, and width dimensions.
% AB - {\color{red} Introduce notation $(T)_{i,j,c}$ for any 3D tensor $T$.}
Given an attacked input $\tilde{\mathbf{x}}$, a defense mechanism based on internal analysis studies one or more deep features to extract a heatmap $\mathcal{H} \in \mathbb{R}^{H^l \times W^l}$, which can in turn allow to highlight the position of the adversarial object within the input image. 

Then, the heatmap can then be binarized, using a threshold, to obtain a mask $\mathcal{M_\delta} \in \{0,1\}^{H^l \times W^l}$, where the elements set to $0$ are deemed adversarial while those set to $1$ are not. 
% GR - Ho chiarito i valori della maschera qua. 
Formally speaking, these steps can be summarized by means of a function $\Lambda^\xi:  \mathbb{R}^{C^l \times H^l \times W^l} \to \{0,1\}^{H^l \times W^l} $, which takes as input the features $\bm{h}^{l}$ from a given layer and produces a mask based on a pre-determined threshold $\xi$.
The resulting mask can then be applied at any layer $L^z$ (e.g., even the input image itself, $z=0$) to filter out the adversarial object, thereby aiming at making the attack ineffective, i.e.,
\begin{equation}
    f^{z \to N_L}( f^{0 \to z}(\tilde{\mathbf{x}}) \odot r^{l \to z}(\mathcal{M_\delta})) \approx f(\mathbf{x}),
\end{equation}
where $\odot$ is the Hadamard product operator on the spatial dimensions and $r^{l \to z}$ is a resizing function to apply a mask extracted at the $l$-th layer to the $z$-th layer (clearly not needed when $l=z$). 
In general, function $r^{l \to z}$ consists of a simple interpolation.
Finally, it is also convenient to define the complementary mask $\mathcal{\bar{M}_{\delta}} = \mathds{1} - \mathcal{M_\delta}$. 
%In general, In some cases, a resizing function may need to be applied to the mask whenever the addressed layer refers to a different spatial dimension. This issue is generally addressed by performing a simple interpolation.

Since this work proposes a defense mechanism for multi-frame cases, from Section~\ref{sec:adv_multiple} on we adopt a discrete-time notation with the superscript $k$ to refer to the symbols introduced above when related to the $k$-th frame, where $k \in \{0, \ldots, K\}$.

\section{Adversarial-Channel Attention}
Inspired by prior research on RW attacks and internal over-activation analysis for neural networks, 
we observed that attacks can be detected by even analyzing shallow network layers only (as opposed to deep layers as done by previous work).
In the following, we provide insights into the existence of over-activation patterns within shallow layers and then address the definition of \emph{adversarial trace}, which is later used to enable the implementation of an adversarial attention mechanism for multi-frame scenarios.

\subsection{Single-frame Adversarial Attention Analysis}
We start by providing insights that link abnormal activations induced by adversarial objects with a particular pattern of channels in the shallow layers.
%This connection primarily arises from the behavior of trained models, which enforce hierarchical feature relationships in neural networks \cite{zeiler2014visualizing}.
%This connection primarily arises from the nature of trained models, which inherently enforce hierarchical feature relationships in neural networks \cite{zeiler2014visualizing}.

\begin{figure}[ht]
\begin{center}
\includegraphics[width=\columnwidth]{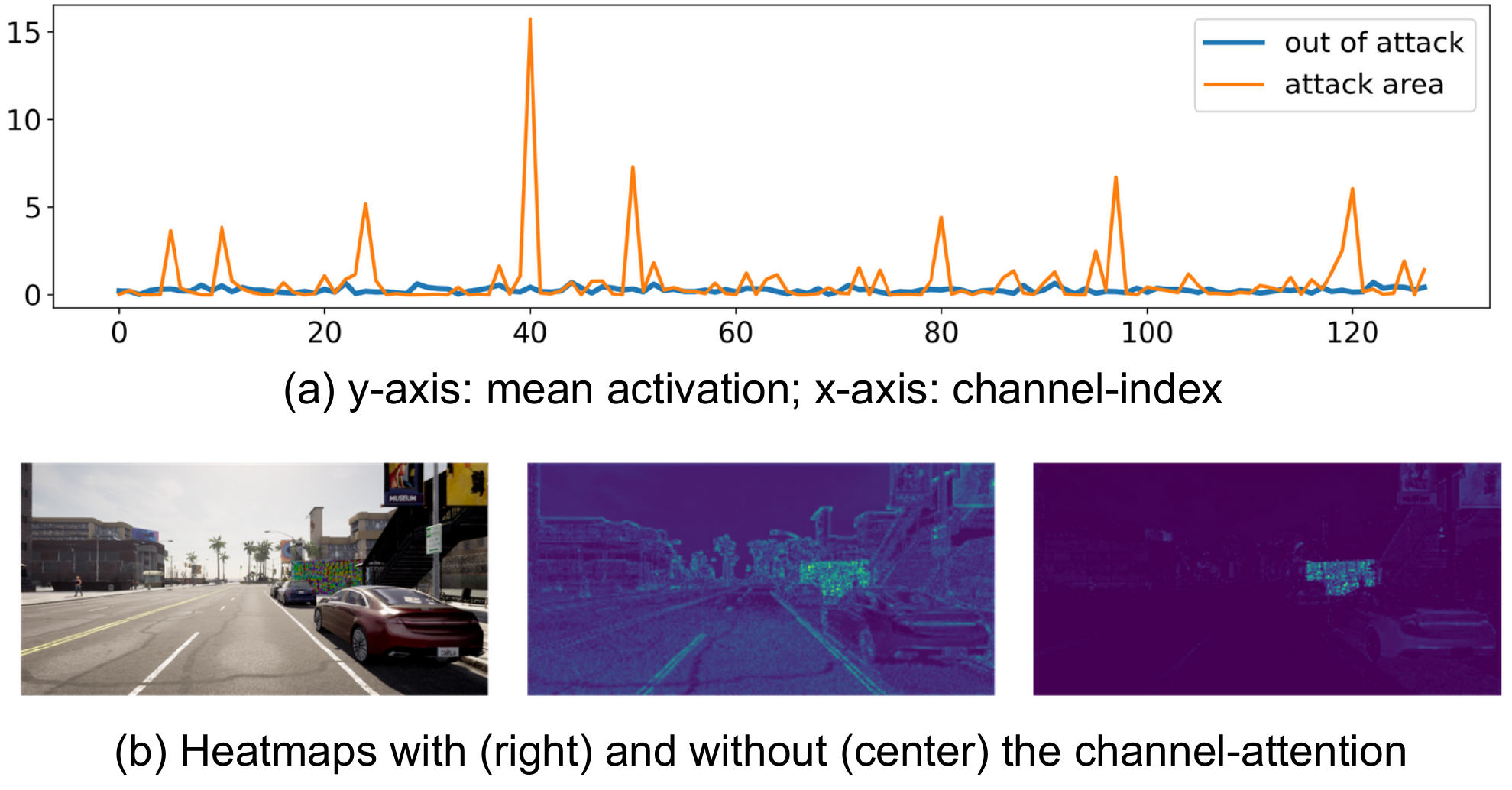}
\caption{\small{(a) Mean channel-wise activation from the first spatial BiSeNet \cite{bisenet_paper} layer during the inference of the attacked image; (b) representation of the heatmap w/ and w/o the attention mechanism.}}
\label{f:sec3}
\end{center}
\end{figure}

\begin{oss}
An adversarial object $\delta$ is designed to minimize a specific adversarial loss function by influencing certain network features (see Eq.\eqref{eq:adv}). %~\cite{yu2021defending}, \cite{rossolini2023defending}. 
%which are in turn linked to the robustness of the attack against different backgrounds and samples.
%This observation can be formalized as follows.
As for instance shown in Figure \ref{f:sec3}(a), we argue that in any layer $L^l$ with activations $\bm{h}^{l}$, there exists a subset of channels targeted by the adversarial object. The channels can be identified by leveraging some channel weights $\boldsymbol{\sigma} \in [0,1]^{C^l}$ that, if applied to $\bm{h}^{l}$, amplify the adversarial features, i.e.,
%if the activations $\bm{h}^{l}$ are channel-wise weighted with $\boldsymbol{\sigma}$, the adversarial effect is consistently more pronounced, that is:
\[
\mathcal{L}_{\text{Adv}}(f^{l\to N_L}(\boldsymbol{\sigma} \cdot \bm{h}^l), \mathbf{y}_{\text{Adv}})~ \leq ~\mathcal{L}_{\text{Adv}}(f^{l\to N_L}(\bm{h}^l), \mathbf{y}_{\text{Adv}}).
\]
\end{oss}
%compared to uniform averaging across all channels.
\begin{oss}
As known from previous work~\cite{yu2021defending}, the adversarial features introduced by physical attacks are characterized by over-activations.
Therefore, from the perspective of an internal analysis (as introduced in Section \ref{sec:internal_analysis}),
%since a higher adversarial effect is linked to a higher over-activation effect \cite{ yu2021defending, rossolini2023defending, co_real}, 
a proper definition of $\boldsymbol{\sigma}$ also allows focusing on the channels that are more subject to over-activation within the attacked area. This results in a better separation of the attacked area from all the others in the heatmap $\mathcal{H}$, which  can be interpreted as a more accurate computation of defense masks.
Formally, this observation can be formulated by means of the intersection-over-union (IoU)~\cite{iou_survey}, which provides the amount of overlap between two masks.
If $\mathcal{M}_{\textit{GT}}$ is the ground-truth mask capable of perfectly masking 
the attacked area in the input image, this observation can be written as a function of the IoU between the predicted attacked area and the ground-truth mask, i.e.,
\[
\textit{IoU}\left( \Lambda^{\xi^{'}}(\boldsymbol{\sigma} \cdot \bm{h}^l),~\mathcal{M}_{\textit{GT}} \right) \geq \textit{IoU}\left( \Lambda^{\xi^{''}}(\bm{h}^l),~\mathcal{M}_{\textit{GT}} \right),
\]
\noindent
where $\xi^{'}$ and $\xi^{''}$ are two thresholds used to extract the masks from activation values $(\boldsymbol{\sigma} \cdot \bm{h}^l)$ and $(\bm{h}^l)$, respectively.\footnote{Note that the thresholds must be different because $\boldsymbol{\sigma}$ scales $\bm{h}^l$.} 
\end{oss}

Channel weights $\boldsymbol{\sigma}$ play a pivotal role in efficiently identifying over-activated areas associated with adversarial features, even in shallow layers. 
In fact, while the over-activation phenomenon may look straightforward to detect, 
our preliminary experiments revealed that simple operations directly applied to all channels, e.g., a channel-wise sum compared to a threshold, do not allow detecting the presence of adversarial objects.

\subsection{Computing the Adversarial Trace}
\label{sec:adv_multiple}
Following the above observations, we propose a practical usage and update of channel weights $\boldsymbol{\sigma}$, which are used to track an adversarial object over time. As anticipated in the paper introduction (see Figure~\ref{f:fig_intro}), the proposed implementation is conceived to be complemented with a defense method capable of providing a starting mask $\mathcal{M}^{0}_{\bm{\delta}}$. When and how this starting mask needs to be computed will be discussed in Section~\ref{s:ACAT}, where the complete defense framework is presented.

The \emph{adversarial trace} is defined as a sequence of weights that highlight the channels over-activated by adversarial attacks.
Formally, given a layer $L^l$, the adversarial trace at time $k$, denoted by $\boldsymbol{\sigma}^{k}$, is a vector of $C^l$ elements in $[0, 1]$ that enables the computation of an accurate heatmap $\mathcal{H}^k$ at time $k > 0$ as follows:
\begin{equation}
\mathcal{H}^k = \sum_{c = 1}^{C^l} (\boldsymbol{\bm{\sigma}}^{k})^{\tau} \cdot \bm{h}^{l,k},
\label{eq:heatmap}
\end{equation}
\noindent where parameter $\tau$ is introduced to amplify the attention pattern within the heatmap. Figure~\ref{f:sec3}(b) shows the benefits of using attention based on the adversarial trace. 
Once the heatmap is obtained, a threshold parameter $\xi^k$ (defined below) can be used to devise the binary mask $\mathcal{M}^{k}_\delta$. 
% GR - {\color{red}possibly also using a noise filter. GR: toglierei questo}

In this work, we proposed a per-frame update of the adversarial trace, so that the next element for time $k+1$ can be computed as a function of the mask and activations computed at time $k$:
\begin{multline}
\boldsymbol{\sigma}^{k+1} = \\ \mathcal{N}\left(\frac{\sum_{i,j=1}^{H, W}\left(\bm{h}^{l,k} \odot \mathcal{\bar{M}}^k_{\bm{\delta}}\right)_{c,i,j}}{|\mathcal{\bar{M}}^k_{\bm{\delta}}|} -  \frac{ \sum_{i,j=1}^{H, W}\left(\bm{h}^{l,k} \odot {{\mathcal{M}}^k_{\bm{\delta}}}\right)_{c,i,j} } {|{{\mathcal{M}}^k_{\bm{\delta}}} |}\right),
\label{eq:adv_trace}
\end{multline}

where $\mathcal{M}^k_{\bm{\delta}}$ is the predicted mask at time $k$, $\bar{\mathcal{M}}^k_{\bm{\delta}}$ denotes a complementary mask to address all other tensor values not interested by $\mathcal{M}^k_{\bm{\delta}}$, and $\mathcal{N}$ represents a normalization function that scales the values to the $[0, 1]$ range. In our experiments, we implemented $\mathcal{N}$ as a ReLU function followed by a channel-wise min-max normalization.

In particular, the first fractional term in Eq.~\eqref{eq:adv_trace} provides attention to over-activated patterns within the area of the adversarial object, while the second term provides negative attention to activations outside the same.

The effectiveness of using information obtained from the current frame to compute the next adversarial trace element $\boldsymbol{\sigma}^{k+1}$ was verified by means of experiments (see Section~\ref{sec:exp}).
%It is worth observing that Eq.~\eqref{eq:adv_trace} uses information computed at the previous time $k$ to compute the next value of the adversarial trace $\boldsymbol{\sigma}^{k+1}$.
%It is worth observing that $\boldsymbol{\sigma}^{k+1}$ in our proposed method refers to the mask and activations computed at time $k$ because the mask at time $k+1$ is not available yet, as it relies on the use of the adversarial trace at time $k+1$. We assume that during this time interval, the changes in channel weights targeted by adversarial attacks do not change significantly. This assumption is also supported by the results of our experiments (see Section \ref{sec:exp}).

\begin{figure}[ht]
\begin{center}
\includegraphics[width=\columnwidth]{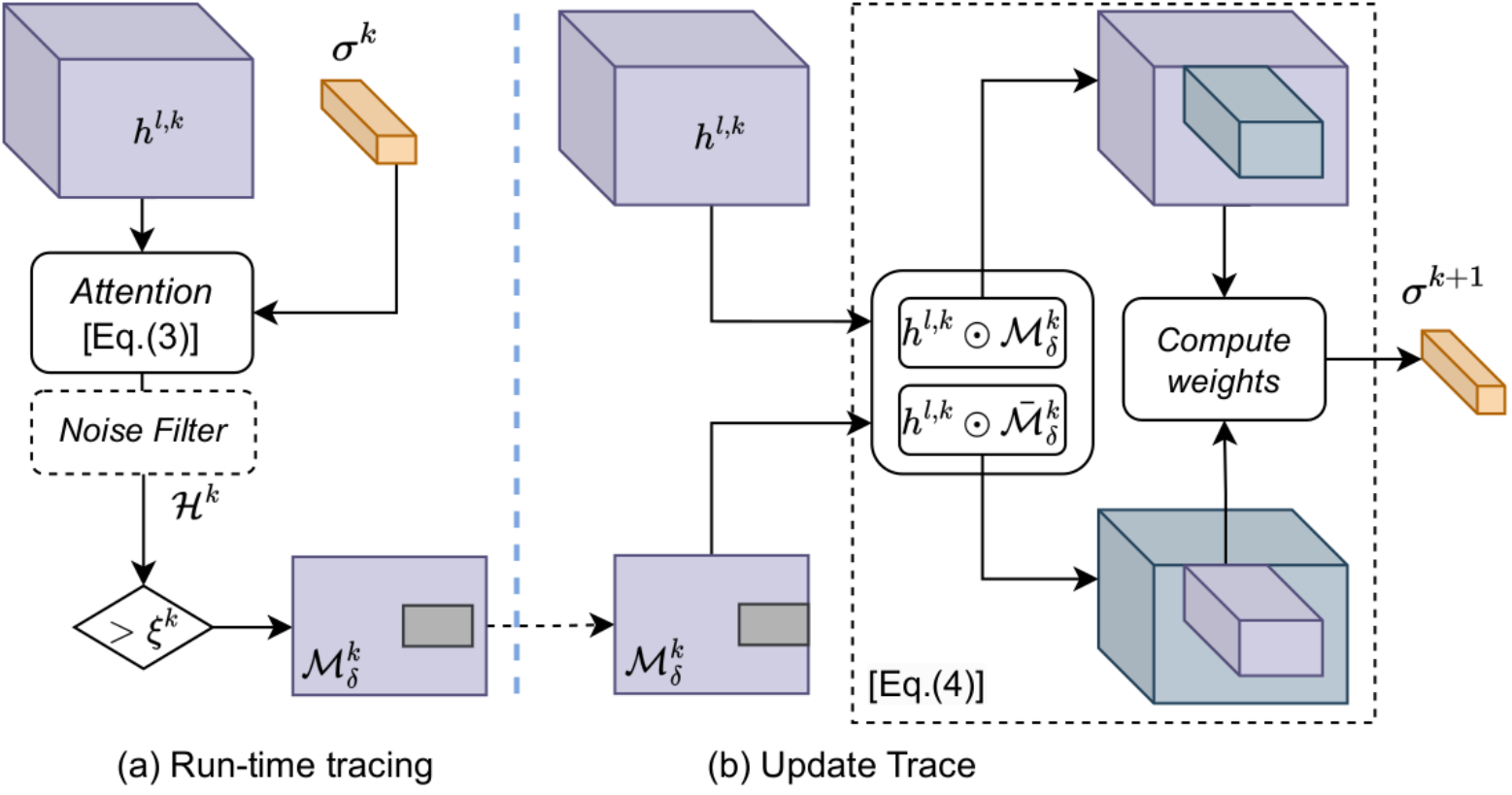}
\caption{\small{Illustration of the operations to implement adversarial attention mechanism performed at time $k$. The resulting output is the next element $\boldsymbol{\sigma}^{k+1}$ of the adversarial trace.}}
\label{f:pl1}
\end{center}
\end{figure}

\noindent\textbf{Summary of the approach.}
Figure~\ref{f:pl1} provides a schematic representation that illustrates the use and update of the adversarial trace for a frame at time $k$. Note that a noise filter (e.g., a Gaussian filter) can be introduced into the pipeline for computing the heatmap. As highlighted in~\cite{rossolini2023defending}, noise filters help mitigate the effects of small spurious activations.

%In summary, the attention mechanism begins with extracting the mask at time $k$ using Equation \ref{eq:heatmap}, followed by updating the adversarial trace for use in frame $k+1$.

\noindent\textbf{Threshold definition.}
Differently from previous work, which adopted a static threshold computed offline on a calibration dataset, this work adopts an adaptive threshold that is dynamically computed  frame by frame. This is necessary due to the attack-specific channel weighting of the attention mechanism, which makes not effective thresholds computed a priori. In ACAT, the threshold is updated at each frame as follows:

%because the attention mechanism weights channels depending on the specific attack, making it hard to compute a threshold in advance. The threshold is computed as follows: 
\begin{equation}
\xi^{k+1} = \max( \mathcal{\bar{H}}^{k} ) + \psi(\mathcal{\bar{H}}^{k}).
\label{eq:threshold-dyn}
\end{equation} 

In the above equation, $\mathcal{\bar{H}}^{k} = \mathcal{H}^{k} \odot D(\mathcal{\bar{M}}^{k}_{\bm{\delta}})$, where $D(\cdot)$ is an operator that expands $\mathcal{\bar{M}}^{k}_{\bm{\delta}}$ by means of an unitary kernel convolution. This expansion is designed to account for \emph{uncertainty} in the areas around the mask, coping with potential spurious activations close to its border. After applying an unitary convolution, non-integer values can be obtained: hence, the operator $D(\cdot)$ eventually binarizes all values using a threshold equal to 0.5.
%converts the values back to a binary form using a threshold of $0.5$ and computes the complementary part, excluding the mask and the uncertainty areas.
%over-activation gradually decreasing from attacked to non-attacked pixels, which could affect threshold values across multiple frames. The introduction of the $D$ operator creates an area of uncertainty not accounted for in the threshold calculation.

To further reduce false positives, an extra safety margin $\psi(\mathcal{\bar{H}}^{k})$ is included in Eq.~\eqref{eq:threshold-dyn}. It is computed as the difference between the $v$-th percentile of the values in $\mathcal{\bar{H}}^{k}$ and the mean value of the same. In our experiments, we used $v=70$, which proved to offer effective resilience to uncertainty. 

 \section{ACAT Framework}\label{s:ACAT}
This section shows how to integrate adversarial-channel attention within the continuous processing loop of vision applications. Algorithm~\ref{alg:algorithm} reports the pseudocode of the operations to be performed at each frame (retrieved with function \textsf{capture\_frame()}). To improve readability, the discrete-time notation with the superscript $k$ is omitted in the pseudocode, as all variables are updated to be used at the next cycle.

\begin{algorithm}
\caption{Adversarial-Channel Attention Tracing}
\label{alg:algorithm}
\begin{algorithmic}[1]

\STATE $\boldsymbol{\sigma} \gets$ None

\WHILE{True}
    \STATE $x \gets$ \textsf{capture\_frame}()
    \IF{$\boldsymbol{\sigma}$ is None}
        \STATE $(y, \mathcal{M}_\delta, \bm{h}^l) \gets$ \textsf{inference\_with\_SoA\_method}($x, f$) \label{line:soa}
        \IF{$\mathcal{M}_\delta$ is not None}
            \STATE {\#\textit{Attack notified}}
            \STATE $\boldsymbol{\sigma} \gets$ \textsf{ACAT\_update}($\bm{h}^l $, $\mathcal{M}_\delta)$ \quad {\#\textit{Eq.~\eqref{eq:adv_trace}}} \label{line:upd1}
            \STATE $\xi \gets$  \textsf{compute\_threshold}($\bm{h}^l $, $\mathcal{M}_\delta)$ \quad {\#\textit{Eq.~\eqref{eq:threshold-dyn}}} 
            \STATE $y = f(x \odot  \mathcal{M}_\delta)$ \quad {\# \textit{Inference with masked input}} \label{line:th1}
            %\STATE $y = f^{[l\rightarrow{L}]}(\bm{h}^l  \odot \mathcal{M}_\delta)$
        \ENDIF
        \STATE \textbf{Continue}  \quad {\#\textit{Wait for next frame}}
    \ENDIF
    \STATE $\bm{h}^l = f^{[0\rightarrow{l}]}(x)$  \label{line:acat-b1}
    \STATE $\mathcal{H} = \mathsf{noise\_filter}\left(\sum_{c = 1}^{C^l} (\boldsymbol{\bm{\sigma}})^{\tau} \cdot \bm{h}^{l}\right)$ \quad {\#\textit{Eq.~\eqref{eq:heatmap}}}
    \STATE $\mathcal{M}_\delta \gets \Lambda^{\xi}(\mathcal{H})$ \quad {\#Apply threshold to get mask} \label{line:acat-e1}
 %   \STATE $\mathcal{M}_\delta \gets$ \textbf{ACAT\_run}($x, \mathcal{M}_\delta, \bm{h}^l, \boldsymbol{\sigma}_{\delta}, \xi$)
    \IF{$|\mathcal{\bar{M}}_\delta| < \lambda_\mathcal{M}$}  \label{line:stop}
        \STATE $\boldsymbol{\sigma} \gets$ None \quad {\#\textit{Stop adv. tracing}}
        \STATE $y = f^{[l\rightarrow{L}]}(\bm{h}^l)$
    \ELSE
        \STATE  $\boldsymbol{\sigma} \gets$ \textsf{ACAT\_update}$(\bm{h}^l,\mathcal{M}_\delta)$ \quad {\#\textit{Eq.~\eqref{eq:adv_trace}}}  \label{line:acat-b2}
        \STATE $\xi \gets$  \textsf{compute\_threshold}($\bm{h}^l $, $\mathcal{M}_\delta)$ \quad {\#\textit{Eq.~\eqref{eq:threshold-dyn}}} \label{line:acat-e2}
        \STATE $y = f^{[l\rightarrow{L}]}(\bm{h}^l \odot \mathcal{M}_\delta)$ {\#\textit{Inference with masked layer}} \label{line:defend}
    \ENDIF
    
\ENDWHILE
\end{algorithmic}
\end{algorithm} 

For each frame, it checks if the adversarial trace $\boldsymbol{\sigma}$ exists. If not, it means no adversarial attack was detected at the previous frame. In this case, a state-of-the-art attack detection method, e.g.,~\cite{rossolini2023defending}, is executed (line~\ref{line:soa}) with a single inference pass. If the latter detects an attack, the algorithm initializes the adversarial trace $\bm{\sigma}$, computes the threshold $\xi$, and leverages the mask compute by the state-of-the-art method to defend from the attack (lines~\ref{line:upd1}-\ref{line:th1}). The processing of the current frame can hence end.

Otherwise, when the adversarial trace $\boldsymbol{\sigma}$ is
available from the previous frame, the algorithm leverages it to compute the defense mask following the results of Sec.~\ref{sec:adv_multiple} (lines~\ref{line:acat-b1}-\ref{line:acat-e1}).
If the mask is meaningful (details provided next), it also computes the next adversarial trace and threshold (lines~\ref{line:acat-b2}-\ref{line:acat-e2}), still based on Sec.~\ref{sec:adv_multiple}, and continues the inference process by applying the mask at the inner layer $L^l$ to defend from the attack (line~\ref{line:defend}).
%set (indicating an ongoing attack), the algorithm executes the ACAT trace on the current frame $\bm
%{x}$, leveraging the previous trace and threshold to compute a new mask. If the mask is still available (line 18, i.e., larger than $\lambda_\mathcal{M}$), it updates the trace and the threshold (line 22 and 23) and filters out the adversarial features, thus forwarding the new activations into the remaining inference part (line 24). 

\paragraph{Reset criterion}
Knowing about the connection between the mask size and the induced adversarial effect by the masked attack~\cite{rossolini2023defending}, we use the number of pixels detected in the predicted complementary mask to decide whether to reset adversarial tracing or not. This could mean that the adversarial object is either too small or far away from the camera. Specifically, we disable adversarial tracing when the computed mask has less than $\lambda_\mathcal{M}$ pixels (line~\ref{line:stop}), where the latter is a configurable parameter.
%Note that users have the flexibility to customize this aspect, allowing the mechanism to remain active but skipping the update part or terminating the tracking process.
%In the tested version, once the stop condition succeeds, we terminate the tracking, waiting for a new detection of the attack from a defense mechanism.

\paragraph{Timing performance}
State-of-the-art approaches require \emph{two} inference passes to defend from adversarial attack while, as it can be noted from Algorithm~\ref{alg:algorithm}, once an attack has been detected at a certain frame, ACAT allows defending from the same with just \emph{one} inference pass (completed in two stages at lines~\ref{line:acat-b1} and~\ref{line:defend}, respectively) for the next frame. This holds until tracing is active, i.e., the reset criterion is not reached. Once a new attack will be detected the same will hold for the next frames, and so on and so forth. Overall, ACAT allows significantly improving the timing performance of the defense mechanism (quantitative results provided in the next section) by halving inference times in general, except for the very first frame in which the attack manifests.  

\section{Experimental evaluation}
\label{sec:exp}
The experimental evaluation is focused on semantic segmentation models designed for autonomous driving, which have recently garnered attention due to the need to address real-world adversarial attacks in outdoor scenarios \cite{rossolini_tnnls_2023, zhang2021evaluating}. Please note however that defense mechanisms based on over-activation also work for different computer vision tasks, where the connection between over-activation and adversarial effect persists \cite{rossolini_tnnls_2023, yu2021defending}.

In the following, we first provide details on the experimental settings. Then, we present and discuss different tests and ablation studies conducted to validate the design and benefits of the proposed defense algorithm.
All the experiments were implemented using PyTorch~\cite{pytorch} on a machine with 8xNVIDIA-A100 GPUs.  
%and assess its benefits in a robust video scenario.

% la metto qua perchè altrimenti va alla pagina successiva
\begin{table*}[htpb]
    \begin{center}
    \begin{tabular}{l|r|r|r|r|r|r|r|r|r|r}
 & \multicolumn{1}{l|}{Scene 1} & \multicolumn{1}{l|}{Scene 2} & \multicolumn{1}{l|}{Scene 3} & \multicolumn{1}{l|}{Scene 4} & \multicolumn{1}{l|}{Scene 5} & \multicolumn{1}{l|}{Scene 6} & \multicolumn{1}{l|}{Scene 7} & \multicolumn{1}{l|}{Scene 8} & \multicolumn{1}{l}{Scene 9} \\
\midrule
No Attack &     \cellcolor[HTML]{f2f2f2}28.35 &    \cellcolor[HTML]{FFFFFF}29.66 &    \cellcolor[HTML]{FFFFFF}28.88 &   \cellcolor[HTML]{f2f2f2}24.71 &    \cellcolor[HTML]{FFFFFF}25.10 &    \cellcolor[HTML]{f2f2f2}26.51 &   \cellcolor[HTML]{f2f2f2}20.56 &    \cellcolor[HTML]{FFFFFF}24.46 &   \cellcolor[HTML]{f2f2f2}23.68 &   \\
\midrule
No Def        &    \cellcolor[HTML]{f2f2f2}-10.2 &    \cellcolor[HTML]{FFFFFF}-4.32 &    \cellcolor[HTML]{FFFFFF}-2.81 &   \cellcolor[HTML]{f2f2f2}-6.4 &    \cellcolor[HTML]{FFFFFF}-5.06 &    \cellcolor[HTML]{f2f2f2}-9.5 &   \cellcolor[HTML]{f2f2f2}-7.00 &    \cellcolor[HTML]{FFFFFF}-2.93 &   \cellcolor[HTML]{f2f2f2}-6.75 &   \\
ACAT\textsubscript{GT}    &    \cellcolor[HTML]{f2f2f2}\textbf{-2.9} [1] &    \cellcolor[HTML]{FFFFFF}\textbf{+0.59} [1] &    \cellcolor[HTML]{FFFFFF}\textbf{-2.67} [1] &   \cellcolor[HTML]{f2f2f2}\textbf{-2.11} [2] &    \cellcolor[HTML]{FFFFFF}\textbf{-0.9} [1] &    \cellcolor[HTML]{f2f2f2}-5.4 [1] &   \cellcolor[HTML]{f2f2f2}\textbf{-2.87} [1] &    \cellcolor[HTML]{FFFFFF}\textbf{-0.63} [1]&   \cellcolor[HTML]{f2f2f2}\textbf{-0.78} [1] &   \\
ACAT\textsubscript{ZM}     &    \cellcolor[HTML]{f2f2f2}\textbf{-2.9} [2] &    \cellcolor[HTML]{FFFFFF}-3.4 [1] &    \cellcolor[HTML]{FFFFFF}-3.15 [12] &   \cellcolor[HTML]{f2f2f2}-2.82 [2] &    \cellcolor[HTML]{FFFFFF}-1.1 [7] &    \cellcolor[HTML]{f2f2f2}\textbf{-5.33} [1] &   \cellcolor[HTML]{f2f2f2}-2.89 [21] &    \cellcolor[HTML]{FFFFFF}-5.92[1]&   \cellcolor[HTML]{f2f2f2}-5.12[1] &   \\
ZMask      &    \cellcolor[HTML]{f2f2f2}-5.32 &    \cellcolor[HTML]{FFFFFF}-2.35 &    \cellcolor[HTML]{FFFFFF}-2.81 &   \cellcolor[HTML]{f2f2f2}-3.01 &    \cellcolor[HTML]{FFFFFF}-1.7 &    \cellcolor[HTML]{f2f2f2}-5.87 &   \cellcolor[HTML]{f2f2f2}-4.84 &    \cellcolor[HTML]{FFFFFF}-3.0 &   \cellcolor[HTML]{f2f2f2}-5.89 &   \\
LGS            &    \cellcolor[HTML]{f2f2f2}-8.63 &    \cellcolor[HTML]{FFFFFF}-5.09 &    \cellcolor[HTML]{FFFFFF}-4.09 &   \cellcolor[HTML]{f2f2f2}-6.43 &    \cellcolor[HTML]{FFFFFF}-4.04 &    \cellcolor[HTML]{f2f2f2}-6.65 &   \cellcolor[HTML]{f2f2f2}-6.01 &    \cellcolor[HTML]{FFFFFF}-1.60 &   \cellcolor[HTML]{f2f2f2}-5.83 &   \\

\midrule
\midrule
No Attack &     \cellcolor[HTML]{FFFFFF}33.04 &    \cellcolor[HTML]{FFFFFF}34.11 &    \cellcolor[HTML]{FFFFFF}34.39 &   \cellcolor[HTML]{f2f2f2}34.22 &    \cellcolor[HTML]{FFFFFF}31.01 &    \cellcolor[HTML]{FFFFFF}33.26 &   \cellcolor[HTML]{f2f2f2}30.73 &    \cellcolor[HTML]{FFFFFF}31.86 &   \cellcolor[HTML]{FFFFFF}37.41 &   \\
\midrule
No Def        &     \cellcolor[HTML]{FFFFFF}+0.65 &    \cellcolor[HTML]{FFFFFF}-1.74 &   \cellcolor[HTML]{FFFFFF}-0.6 &    \cellcolor[HTML]{f2f2f2}-6.53 &    \cellcolor[HTML]{FFFFFF}-1.01 &   \cellcolor[HTML]{FFFFFF}-2.31 &    \cellcolor[HTML]{f2f2f2}-12.69 &   \cellcolor[HTML]{FFFFFF}-3.03 &  \cellcolor[HTML]{FFFFFF} -1.11 &  \\
ACAT\textsubscript{GT}     &    \cellcolor[HTML]{FFFFFF}\textbf{+1.59} [1] &    \cellcolor[HTML]{FFFFFF}\textbf{-1.58} [1] &    \cellcolor[HTML]{FFFFFF}\textbf{+0.66} [3] &   \cellcolor[HTML]{f2f2f2}\textbf{-3.29} [1] &    \cellcolor[HTML]{FFFFFF}\textbf{+0.4} [1] &    \cellcolor[HTML]{FFFFFF}\textbf{-0.83} [1] &   \cellcolor[HTML]{f2f2f2}\textbf{-1.52} [1] &    \cellcolor[HTML]{FFFFFF}\textbf{-1.11} [1]&   \cellcolor[HTML]{FFFFFF}\textbf{-0.86} [1] &   \\
ACAT\textsubscript{ZM}    &    \cellcolor[HTML]{FFFFFF}+0.59 [2] &    \cellcolor[HTML]{FFFFFF}-1.73 [3] &    \cellcolor[HTML]{FFFFFF}-0.67 [15] &   \cellcolor[HTML]{f2f2f2} -3.52 [6] &    \cellcolor[HTML]{FFFFFF}+0.35 [5] &    \cellcolor[HTML]{FFFFFF}-2.85 [1] &   \cellcolor[HTML]{f2f2f2}-4.25 [33] &    \cellcolor[HTML]{FFFFFF}-2.22 [16]&   \cellcolor[HTML]{FFFFFF}-1.01 [5] &   \\
ZMask      &    \cellcolor[HTML]{FFFFFF}-2.27 &    \cellcolor[HTML]{FFFFFF}-3.2 &    \cellcolor[HTML]{FFFFFF}-1.32 &   \cellcolor[HTML]{f2f2f2}-5.2 &    \cellcolor[HTML]{FFFFFF}-0.97 &    \cellcolor[HTML]{FFFFFF}-5.55 &   \cellcolor[HTML]{f2f2f2}-3.58 &    \cellcolor[HTML]{FFFFFF}-0.98 &   \cellcolor[HTML]{FFFFFF}-1.01 &   \\
LGS            &    \cellcolor[HTML]{FFFFFF}+0.45 &    \cellcolor[HTML]{FFFFFF}-3.51 &    \cellcolor[HTML]{FFFFFF}+1.01 &   \cellcolor[HTML]{f2f2f2}-8.30 &    \cellcolor[HTML]{FFFFFF}-0.87 &    \cellcolor[HTML]{FFFFFF}-4.06 &   \cellcolor[HTML]{f2f2f2}-11.97 &    \cellcolor[HTML]
{FFFFFF}-1.22 &   \cellcolor[HTML]{FFFFFF}-1.25 &   \\
\end{tabular}
    \caption{\small{Variation of the multi-class mIoU w/ and w/o defense mechanisms across the 9 driving scenarios of CarlaGear~\cite{nesti2022carla}. Results for both  BiseNet~\cite{bisenet_paper} (top) and DDRNet \cite{ddrnet_paper} (bottom) are reported. The values inside square brackets denote the number of times ACAT required to be re-initialized (reset criterion). The results of ACAT are averaged across 5 random shuffling of each scene dataset.}}
    \label{tab:accuracy-input-size}
    \end{center}
\end{table*}

\subsection{Experimental settings}
Complete multi-frame benchmarks to evaluate the effectiveness of defense methods against real-world adversarial attacks are not available from previous work.

%especially when compared to the more well-explored single-image domain. 
For this reason, we addressed two evaluation approaches:
\textbf{(i)} attack scenarios generated with the CARLA simulator~\cite{carla_alexey}, used to test the attention mechanism of ACAT only, and \\ \textbf{(ii)} digitally attacked video generated with 
Cityscapes~\cite{cordts2016cityscapes}, which instead allow testing the whole ACAT framework.

\paragraph{Attacks in CARLA-simulated scenarios}  With the intent of facing with realistic settings, we utilized the Carla-Gear framework~\cite{nesti2022carla}, which offers 9 photo-realistic scenarios (50 test images each) collected in areas of Carla-town 10~\cite{carla_alexey}, integrating adversarial billboards specifically designed for each model in use. Please note that the framework only provides random viewpoints of the area next to the adversarial billboards, which are not sequential videos.
For this reason, this setting allows evaluating the benefits of adopting adversarial-channel attention only, i.e., improving the capabilities of state-of-the-art defense mechanisms when used on a single frame, while not enabling meaningful tests to evaluate ACAT as a whole.
%Although the framework only provides random viewpoints of the area close to the billboard, which are not sequential videos, it allows showing the benefits of using adversarial attention.
%beyond a multi-frame application.

\paragraph{Digitally attacked video datasets} To address the lack of a dedicated video dataset featuring attacked driving scenes, we generated custom videos that include digital adversarial attacks. Three extended sequences from Cityscapes~\cite{cordts2016cityscapes} videos\footnote{https://www.cityscapes-dataset.com, \textit{leftImg8bit\_demoVideo.zip}} were utilized with images sized at 2048x1024 pixels. Within each video, a dynamic adversarial patch was digitally introduced in the frames, which, at every frame, changes its position and scaling factor, following a sinusoidal trend.
The patch position and scale were computed as follows:
\begin{equation}
 \begin{bmatrix}
  x_{\textit{pos}} \\[0.5em]
  y_{\textit{pos}} \\[0.5em]
  s
\end{bmatrix}
=
\begin{bmatrix}
  c_x + A_x \sin{(\alpha_x \cdot k + \omega_x)} \\[0.5em]
  c_y + A_y \sin{(\alpha_y \cdot k + \omega_y)} \\[0.5em]
  1 + A_s \sin{(\alpha_s \cdot k + \omega_s)}
\end{bmatrix},
\label{eqn:patchmove}
\end{equation}
\noindent where $k$ is the frame index, $x_{\textit{pos}}$ and $ y_{\textit{pos}}$ are coordinates of the position of the patch, $c_x, c_y$ are the center coordinates of the frame, and $s$ is the scaling factor of the patch. In our experiments we set $(A_x, A_y, A_s, \alpha_x, \alpha_y, \alpha_s) = (500, 300, 0.3,$ $0.05, 0.05, 0.05)$. The $\omega$ values represent a phase used to randomize tests among different initial positions.
With these settings, the patch can partially go beyond the image boundaries while holding a size that is sufficient for producing an adversarial effect \cite{rossolini_tnnls_2023, patch_object}. The $\alpha$ values provide a smooth trend of the patch among subsequent frames.
%Further illustration and implementation details of the digital attacked video Cityscapes datasets are available in the supplementary material.

The attack mechanism used to generate the patch was the Over-Activation-aware Expectation Over Transformation (EOT) optimization~\cite{rossolini2023defending, pmlr-v80-athalye18b}, where a parameter $\beta \in [0,1]$ is used to regulate the over-activation level of the patch within the internal layers  while reducing the adversarial effect (the lower the $\beta$ the lower the over-activation, and so the adversarial effect). 
%In our test, we noticed that an acceptable adversarial effect is available for $\beta > 0.6$. 
%Opting for medium values of $\beta$ (e.g., $0.6$) enables the creation of more constrained attack patches that maintain their adversarial aspects while controlling the level of over-activation, closely resembling a real-world scenario.
This approach is particularly useful for evaluating the robustness of our approach when the attacker tries to limit over-activation to mount attacks that are difficult to detect. 

%$\text{Mask-IoU} =\frac{1}{|X|} \sum_{\tilde{\bm{x}} \in \tilde{X}} \textit{IoU}(\mathcal{M}_{\bm{\delta}(\bm{x})},  \mathcal{M}_{\textit{gt}}(\bm{x}))$
% \[
% \text{Mask-IoU} =\frac{1}{|X|} \sum_{x}\frac{\sum_{i,j}^{H, W} \mathcal{M}^k_{\bm{\delta}} \odot \mathcal{M}^k_{\textit{gt}}}{\sum_{i,j}^{H, W} \mathcal{M}^k_{\bm{\delta}} +\mathcal{M}^k_{\textit{gt}} - (\mathcal{M}^k_{\bm{\delta}} \odot \mathcal{M}^k_{\textit{gt}})},
% \]
%\noindent where $\mathcal{M}^k_{\textit{gt}}$ is the ground truth mask of the patch and $\tilde{X}$ is the set of attacked frames.  

\paragraph{Network models and defense methods}
Following related work on semantic segmentation~\cite{nesti2022carla}, we considered real-time high-performance DNN models: DDRNet-Slim23 version~\cite{ddrnet_paper} and BiSeNetX39~\cite{bisenet_paper}. We use the pre-trained versions available from~\cite{nesti2022carla}.

We compared our method ACAT with two lightweight single-frame approaches designed to mask real-world attacks. The first is LGS~\cite{naseer_local_2019}, which applies gradient-based filtering of the image to mask adversarial pixels. The second is ZMask~\cite{rossolini2023defending}, which, as for ACAT, is based on over-activation but necessitates of two forward passes at any frame, since addresses also deep network layers. For both, we utilized the original settings provided by the respective authors.

For ACAT, we set $\tau = 2$, and the kernel size to $5,3$ and $31, 11$ for the Gaussian filter and dilatation operators for Bisenet and DDRNet, respectively. The different sizes are due to the different spatial dimensions of the features.
The layers analyzed by ACAT are in the shallower blocks of the considered model, specifically the output of the second block of DDRNet and the output of the first context layer of BiSeNet. Ablation studies were also performed to understand the selection process of these layers (see Sec.~\ref{s:layer_ablation}).

\paragraph{Metrics}
Different metrics were used to assess the performance of the addressed mechanisms. Given the unavailability of annotations for the Cityscapes videos, 
we use the binary Intersection-over-Union (IoU), referred to as Mask-IoU, to measure the overlap between the predicted complementary mask $\mathcal{\bar{M}}_{\bm{\delta}}^k$ (whose values equal to $1$ denote the predicted adversarial region) and the corresponding ground-truth mask $\mathcal{\bar{M}}^k_{\textit{GT}}$. Intuitively, Mask-IoU quantifies the quality of the predicted defense mask: the higher the better.
% AB - comment {Intuitively,\color{red} need to give intuition about what it is.}

For the tests conducted on the Carla-Gear dataset, as indicated in the benchmark, we measured the effectiveness of adversarial attacks by addressing the original multi-class MIoU \cite{nesti2022carla, cordts2016cityscapes} of the task, since annotations are available.
\subsection{Performance Evaluation on Carla} 
Table~\ref{tab:accuracy-input-size} highlights the advantages of our approach across nine scenarios of the Carla-Gear dataset on BiSeNet (top part) and DDRnet (bottom). Regarding ACAT, which is designed to integrate with state-of-the-art defenses, we conducted analyses under two settings: $\textit{ACAT}_{\textit{ZM}}$ and $\textit{ACAT}_{\textit{GT}}$. The former utilizes ZMask~\cite{rossolini2023defending}, reflecting a realistic scenario built upon an already available approach. In the second setting, $\textit{ACAT}_{\textit{GT}}$ assumes the knowledge of an ideal, ground-truth mask at first frame in which the attack is detected. While this setting depicts a less realistic scenario, it serves to highlight the intrinsic performance of ACAT, independently from the defense method with which it is integrated.

In the table, the first line for each scenario depicts the task MIoU without an adversarial billboard, while subsequent lines show the drop in MIoU with the adversarial billboard and/or without the related defenses. The value between the brackets for the ACAT results depicts the number of times that the reset criterion takes effect, necessitating the extraction of a new starting mask.
As also mentioned in~\cite{nesti2022carla}, there are instances where certain attacks can be particularly challenging for a specific model and scenario, leading to a poor reduction in the MIoU. To assist the reader, in Table~\ref{tab:accuracy-input-size} we highlighted in gray the scenarios that have resulted in a more pronounced adversarial effect.

As it can be noted from the table, ACAT consistently outperforms the other methods, significantly reducing the number of extra inference passes, reaching the reset conditions only a few times. Note also that $\textit{ACAT}_{\textit{ZM}}$ generally improves the performance of ZMask. However, when ZMask fails to return an accurate mask, it may jeopardize the initialization of ACAT, resulting in lower performance (e.g., scene 8 - BiSeNet and scene 7 on DDRNet). This is not the case for $\textit{ACAT}_{\textit{GT}}$, confirming that the lower performance is not due to ACAT.
%showing that this kind of issues are likely due to ZMask only, rather than the ACAT algorithm. 
Concerning LGS, as known from previous work, it loses accuracy in real-world scenarios \cite{rossolini2023defending, chiang_adversarial_2021}.

Please note that, in these tests only, we did not update the trace and threshold of ACAT (lines~\ref{line:acat-b2}-\ref{line:acat-e2} in Algorithm \ref{alg:algorithm}). As anticipated above, this is because the tested images do not pertain to sequential video. 
The whole ACAT framework is instead addressed by the following experiments.

It is however interesting to also observe the number of times ACAT required a re-initialization (reset criterion) during these tests, even if updates are disabled. As one may expect, we found scenarios in which the mask provided by ZMask was frequently required (e.g., note the numbers between square brackets in Table~\ref{tab:accuracy-input-size} for Scenes 3 and 7), while surprisingly, in other cases, it was not at all. This means that the attention mechanism offered by ACAT is sometimes effective even with sporadic updates (see also the other experiments below). Conversely, in the former case, we found that the reset criterion was prominently triggered because the mask provided by ZMask was not particularly accurate, as $\textit{ACAT}_{\textit{GT}}$ almost never requires to be re-initialized.

\subsection{Performance Evaluation on Digital Attacks}
\begin{figure}[ht]
\begin{center}
\includegraphics[width=\columnwidth]{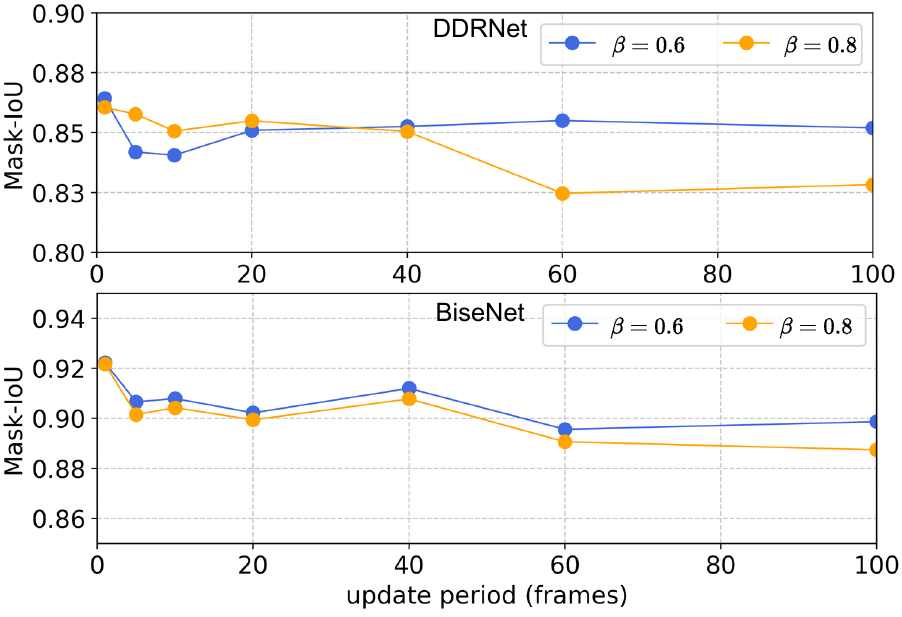}
\caption{\small{Mask-IoU performance by varying the update period (in frames, x-axis of the figure) of the adversarial trace and thresholds. The upper plot refers to the DDRNet architecture, while the second one pertains to BiseNet. The tests evaluate the performance of $\textit{ACAT}_\textit{ZM}$ for two distinct digital patches ($\beta = 0.6$ and $\beta = 0.8$). The results are the average of five different initializations of the $\omega$ parameters in Eq.~\eqref{eqn:patchmove}.}}
\label{fig:change_frames}
\end{center}
\end{figure}

In Figure~\ref{fig:change_frames}, we studied the Mask-IoU for the digital attacked video. To show that ACAT provides high robustness even when the adversarial trace and thresholds are not updated at every frame as mandated by Alg.~\ref{alg:algorithm}, we measured the average Mask-IoU under $\textit{ACAT}_{\textit{ZM}}$ on attacked video streams from Cityscapes, varying the period with which the trace and thresholds are updated.
The period is expressed in number of frames and is reported on the x-axis of the figure (e.g., value 1 on the x-axis means that the update occurs at each frame). In the analysis, we tested two digital adversarial patches, with $\beta = 0.6$ and $\beta = 0.8$, to better investigate on the robustness of ACAT.
%The figure also reports the results for ZMask (dashed lines), which are constant because ZMask is a single-frame mechanism.
We also evaluated ZMask, which achieves $(0.66, 0.75)$ and $(0.70, 0.72)$ of Mask-IoU with ($\beta$ = 0.6, $\beta$ = 0.8) for Bisenet and DDRNet, respectively.
The results for LGS are not reported since it does not provide a binary defense mask, but rather a soft filtering of the input image, for which it is not possible to compute the Mask-IoU.

The figure shows that ACAT surprisingly works well even with sporadic updates of the adversarial trace and thresholds. This was also due to the fact that the Cityscapes videos are related to rather static scenarios. 
In fact, despite some changes in the appearance of adversarial patches and their background, the over-activated pattern of the patch in these cases continuously insist on a similar set of channels to induce the adversarial effect. %This suggests that the algorithm can maintain its effectiveness even with rare frequent updates.
An update of the parameters is anyway required in more dynamic scenarios with more frequent changes of the background and appearance of the adversarial object. 
%As shown in the figure, the approach produces good results even with a long update time period. This resilience can be 
The figure also shows that sporadic updates always provide better performance than ZMask.

%Concerning comparisons with ZMask, we can see that the analysis performed in shallow layers provide a more accurate Mask-IoU, even concerning more critical attacks with lower values of $\beta$.

\begin{comment}

\end{comment}

\subsection{Ablation Studies}
To better understand the contribution of each operation performed by ACAT to its overall performance, Table~\ref{tab:ablation} reports the Mask-IoU of $\textit{ACAT}_{GT}$ on the attacked videos under different settings.
With the aim of acquiring a deeper understanding about the attention mechanism of ACAT, we independently examined the two fractional terms defined in Equation~\eqref{eq:adv_trace} to update the adversarial trace. The first term provides positive attention within the attacked area, which is the most important part of the attention mechanism. We hence introduce a flag $\textit{Att}^+$ to indicate a setting of  ACAT that uses this term. The second term refines the previous operation by introducing negative attention to the elements outside the attacked area. Another flag $\textit{Att}^-$ is also introduced to denote if this second term is used by ACAT.

\begin{figure}[ht]
\begin{center}
\includegraphics[width=\columnwidth]{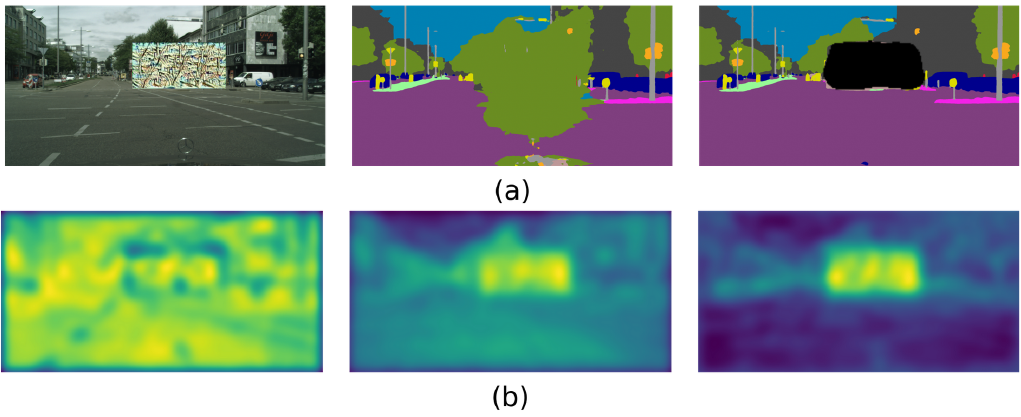}
\caption{\small{(a) Comparison of the adversarial effect of a patch with $\beta = 0.6$ (left), with $\text{ACAT}_\text{GT}$ mechanism (right), and without the $\text{ACAT}_\text{GT}$ mechanism (middle). (b) Illustration of the heatmap among different settings, from left to right: \textbf{(i)} only NF enabled, \textbf{(ii)} only $\textit{Att}^+$ and NF enabled, \textbf{(iii)} $\textit{Att}^+$, $\textit{Att}^-$, NF, and \textit{Upd} enabled.}}
\label{f:ablation_ill}
\end{center}
\end{figure}

As shown in the table, it is clear that using both $\textit{Att}^+$ and $\textit{Att}^-$ leads to better results, hence motivating the construction of Equation~\eqref{eq:adv_trace}.
In general, it is evident that the use of the attention mechanism significantly improves the Mask-IoU when compared to not using attention (both $\textit{Att}^+$ and $\textit{Att}^-$ disabled, first rows of the table). Its benefits are especially notable in the results obtained with DDRNet, where adversarial over-activations in the shallow layers proved to be very difficult to detect without attention. These observations are also illustrated with an example frame in Figure~\ref{f:ablation_ill}.

The ablation studies also tested ACAT with and without the update of the adversarial trace and the threshold of Eq.~\eqref{eq:threshold-dyn} (flag \textit{Upd} in Table~\ref{tab:ablation}), and with and without the noise filter (flag NF).
% To better understand the contribution of each operation performed by ACAT to its overall performance, Table~\ref{tab:ablation} reports the Mask-IoU of $\textit{ACAT}_{GT}$ under different settings. In particular, to assess the advantages of the proposed channel-attention method, we valuated a simpler version of ACAT that only considers the features within the patch area (denoted by CA* in table). This version does not account for negative attention to the areas outside the mask and it is obtained by using the first term of Equation~\eqref{eq:adv_trace} only, i.e., $\frac{\sum_{i,j=1}^{H, W}(\bm{h}^{l,k} \odot \mathcal{M}_{\bm{\delta}})_{i,j,c}}{|\mathcal{M}_{\bm{\delta}}|}$. 
% {\color{red} What's CA without *?}
% The proposed mechanism consistently demonstrates benefits from attention, particularly when the activations in the shallow layers are not explicitly evident (as observed for DDRNet).

% In Figure \ref{f:ablation_ill}, we present an illustration of the effects of different settings within one attacked frame with DDRNet. It is clear that using the channel attention mechanism significantly enhances the performance in extracting better masks in the initial layers.
%
\begin{table}[ht]
    \small
    \centering
    \begin{tabular}{|c|c|c|c|c|c|c|c|c}
    & & & &  \multicolumn{2}{c|}{Bisenet} & \multicolumn{2}{c|}{DDRNet} \\
     $\textit{Att}^+$ & $\textit{Att}^-$  & \textit{Upd} & NF & $\delta_{0.6}$ & $\delta_{0.8}$ & $\delta_{0.6}$ & $\delta_{0.8}$ \\
     \hline
     &  &   &  & 11.9 & 16.2 & 0.00 & 0.01 \\
     \hline
     &  &   &   \checkmark & 89.0 & 88.7 & 7.2 & 0.6  \\
     \hline
    \checkmark & & & \checkmark & 90.7 & 90.2 & 76.91 & 84.90 \\
    \hline
    \checkmark & &\checkmark & \checkmark & 91.3 & 90.8 & 72.05 & 83.05 \\
    \hline
    \checkmark & \checkmark & & \checkmark & \textbf{92.24} &  91.9 & 85.56  & 84.22 \\
    \hline
    \checkmark & \checkmark & \checkmark & \checkmark & 92.23 & \textbf{92.18} & \textbf{86.12} & \textbf{86.42} \\
    \end{tabular}
    \caption{\small{Experimental results of ablation studies with respect to the different components used to update the adversarial trace. The results are in terms of Mask-IoU and related to the digitally-attacked Cityscape videos using $\textit{ACAT}_\text{GT}$ as a defense mechanism. Two model-specific patches were utilized, one with $\beta=0.6$ and another with $\beta=0.8$.}. In the table, $\textit{Att}^+$ and $\textit{Att}^-$ denote two flags to enable the two attention terms of Equation~\ref{eq:adv_trace}, respectively, while \textit{Upd} and the NF are other two flags to enable the update of the trace and threshold, and the usage of the noise filter, respectively.}
    \label{tab:ablation}
\end{table}

\subsection{Layer-wise Ablation} 
\label{s:layer_ablation}
\begin{figure}[htb!]
\begin{center}
\includegraphics[width=\columnwidth]{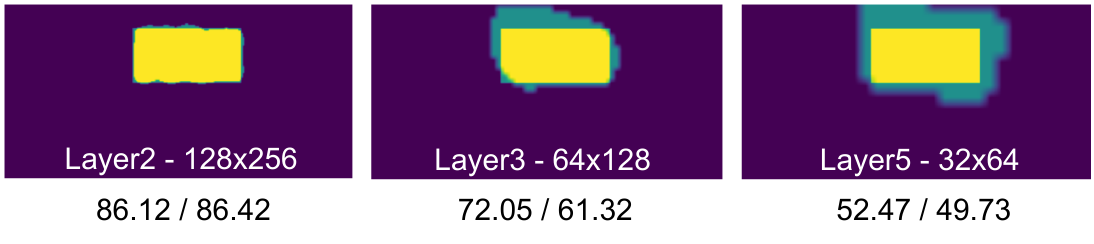}
\caption{\small{Mask-IoU (in black) for the digital adversarial patch with $\beta=0.6$ and $0.8$  on attacked cityscapes video using the $\textit{ACAT}_{GT}$ on different layers of DDRNet. The figures show the overlapping between the predicted mask and the ground truth for $\beta=0.6$, with the highest color indicating the degree of overlap. The depth of the DDRNet layer and the spatial dimension are denoted in white.}}
\label{f:ablation_layers}
\end{center}
\end{figure}

Figure~\ref{f:ablation_layers} reports the Mask-IoU by varying the layer of the DDRNet model with which ACAT operates (parameter $l$ in Alg.~\ref{alg:algorithm}). As observed, the more shallow the layer the better the performance. 
In fact, if addressing deeper layers, the mask based on the over-activation extends beyond the ground-truth position (in the figure, only the yellow parts denote a complete overlap of the ground-truth and the predicted mask). 
This is attributed to the fact that the features of shallow layers are less spatially compressed (i.e., they have a higher spatial size) than those in deeper layers.
%This trend is attributed to (i) the spatial dimension, and (ii) the spreads of adversarial features.

Note that, for fair comparisons, in layer $l=3$, we used the same kernel size as layer $l=2$ (i.e., 3), which provided better performance than kernel size 1 (i.e., no Gaussian filter). While, for layer $l=5$, we did not use the Gaussian filter due to the high compression of the spatial dimension.
These results highlight how ACAT allows focusing on shallow layers so that attacks can be masked within a single inference pass, as opposed to previous work that analyzes deep layers and hence requires another inference pass to mask attacks. % enabling online masking mechanisms within the same model inference.

\subsection{Timing Evaluation}
\begin{figure}[htb!]
\begin{center}
\includegraphics[width=\columnwidth]{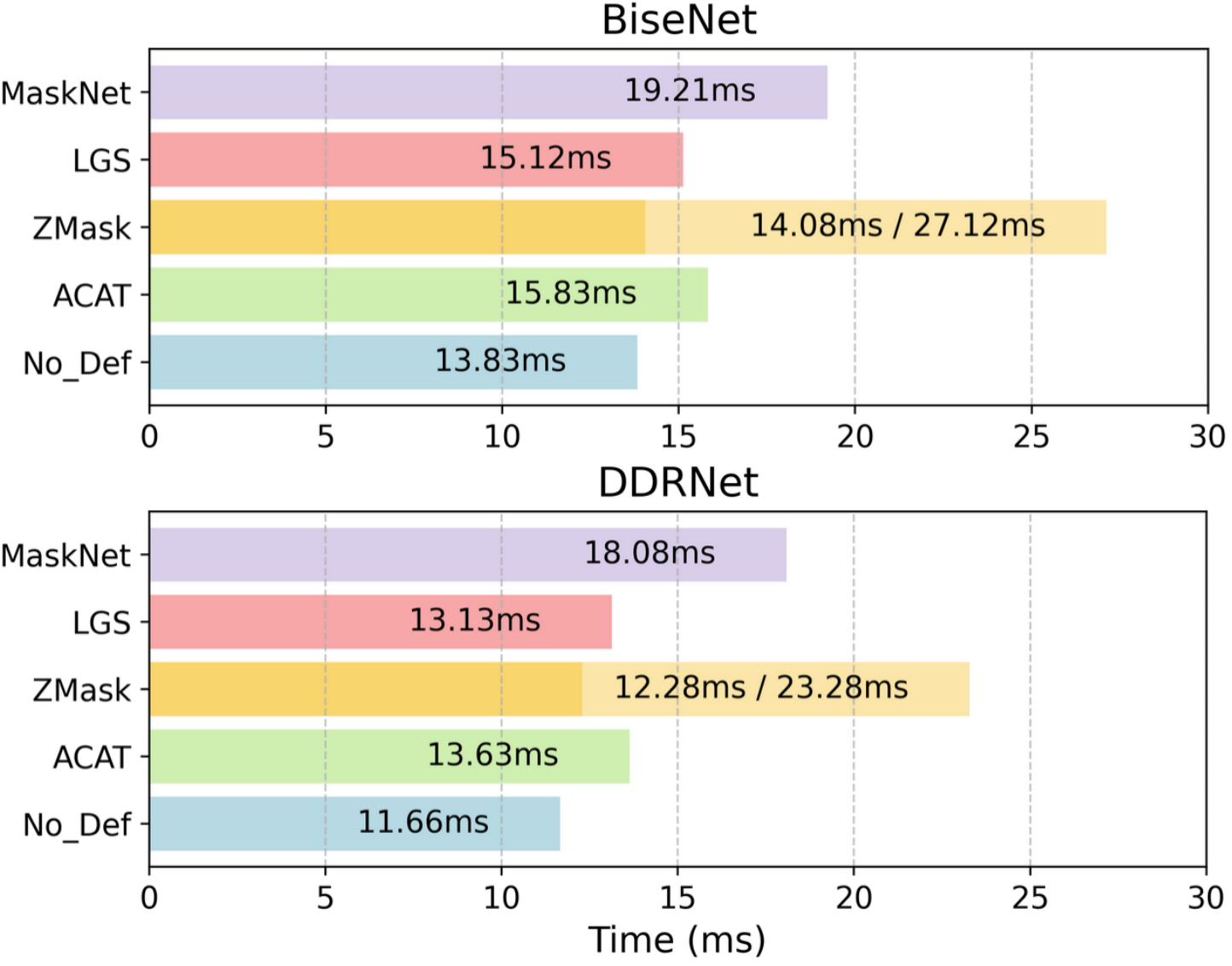}
\caption{\small{Overall inference time with and without defense mechanisms for DDRNet and BiseNet.
}}
\label{f:timing_analysis}
\end{center}
\end{figure}

To demonstrate the improvements provided by ACAT in terms of running times, we measured the inference times when testing the attacked Cityscapes videos. 
Figure~\ref{f:timing_analysis} reports the overall inference time required on average to process a frame by the tested defense mechanisms, with the baseline labeled by \textit{No\_Def}, denoting the original model without defenses. Two inference times are reported for ZMask: when an attack is not detected and when an attack is detected, which are separated by a slash in the figure. As expected, when ZMask detects an attack, its inference time is approximately twice the one of the baseline model. 
Conversely, when no adversarial attacks are detected, ZMask is particularly efficient and hence represents an excellent choice to work in conjunction with ACAT, which activates only when an attack is first detected (see Alg.~\ref{alg:algorithm}).
%demonstrates its efficiency in acting as a simple monitoring mechanism and resulting in a negligible overhead. 

The figure also reports the results for another state-of-the-art defense mechanism, named MaskNet~\cite{chiang_adversarial_2021}, which incorporates a secondary model. It is relatively more expensive due to the necessity of always running an encoder-decoder model in tandem with the original model.

Note that LGS exhibits comparable timing performance with respect to ACAT, since it focuses on specific filters that are directly applied to the input image. However, as shown by the results in Table 1 and other studies in previous work~\cite{chiang_adversarial_2021, rossolini2023defending}, LGS tends not to perform well in detecting adversarial attacks that can be carried out in real world, i.e., by means of physical adversarial objects.

In summary, these results remark on how ACAT provides a well-balanced trade-off between defense performance and overall inference time.

\section{Conclusion}
This work established a novel understanding of the feature over-activation induced by physical adversarial objects in modern neural networks. Differently from previous work, this work proposed an approach that allow identifying physical adversarial attacks by analyzing the first layers of the network, enabling the implementation of efficient defenses for multi-frame vision applications that mostly require just an inference pass to inhibit attacks.
Based on these findings, we proposed \textit{Adversarial-Channel Attention Tracing} (ACAT), a framework based on the concept of \emph{adversarial trace} that focuses on specific channels (within a given layer) that are primarily responsible for propagating the adversarial effect. ACAT is used to extend single-frame defense mechanisms from previous work, which instead may require two inference passes to defend from attacks.

Experimental results demonstrated that ACAT allows both improving the defense capabilities of state-of-the-art defense methods, even when used for a single frame, as well as providing a lower computational cost by detecting and defending attacks in a single inference pass.

Future work should aim at providing a more comprehensive integration of the approach into complex AI-based vision systems. We believe that, beyond the presentation of the ACAT framework, our findings and analyses will also contribute to gaining a deeper understanding of the nature of these physical attacks and, consequently, the development of even more effective defense strategies.
\label{sec:conc}

\balance
\bibliographystyle{IEEEtran}
\bibliography{main}

\end{document}